\def\year{2019}\relax
\documentclass[letterpaper]{article} %DO NOT CHANGE THIS
\usepackage{aaai19}  %Required
\usepackage{times}  %Required
\usepackage{helvet}  %Required
\usepackage{courier}  %Required
\usepackage{url}  %Required
\usepackage{graphicx}  %Required
\usepackage{arydshln}

\usepackage{amssymb}
\usepackage[linesnumbered,ruled]{algorithm2e}
\usepackage{multirow}     
\usepackage{multicol} 
\usepackage{booktabs}
\usepackage{mathtools}

\newcommand{\citet}[1]
{\citeauthor{#1}~\shortcite{#1}}
\newcommand{\citep}{\cite}

\DeclareMathOperator{\RR}{\mathbb{R}}
\DeclareMathOperator*{\id}{\mathbf{I}}

\title{Unsupervised Post-processing of Word Vectors via Conceptor Negation}
\author{Tianlin Liu\thanks{Research done while visiting University of Pennsylvania.} \\
  Department of Computer Science and \\ Electrical Engineering\\
Jacobs University Bremen \\
28759 Bremen, Germany\\
\texttt{t.liu@jacobs-university.de} \\
  \\\And
  Lyle Ungar \and Jo\~{a}o Sedoc\\
Department of Computer and \\ Information Science\\
University of Pennsylvania\\
Philadelphia, PA 19104\\
\texttt{\{ungar, joao\}@cis.upenn.edu}}

\begin{document}
\maketitle
\begin{abstract}

Word vectors are at the core of many natural language processing tasks. Recently, there has been interest in post-processing word vectors to enrich their semantic information. In this paper, we introduce a novel word vector post-processing technique based on \emph{matrix conceptors} \citep{Jaeger2014}, a family of regularized identity maps. More concretely, we propose to use conceptors to suppress those latent features of word vectors having high variances. The proposed method is purely unsupervised: it does not rely on any corpus or external linguistic database. We evaluate the post-processed word vectors on a battery of intrinsic lexical evaluation tasks, showing that the proposed method consistently outperforms existing state-of-the-art alternatives. We also show that post-processed word vectors can be used for the downstream natural language processing task of dialogue state tracking, yielding improved results in different dialogue domains.
\end{abstract}

\section{Introduction}

Distributional representations of words, better known as word vectors, are a cornerstone of practical natural language processing (NLP). Examples of word vectors include Word2Vec \citep{Mikolov2013}, GloVe \citep{Pennington2014}, Eigenwords \citep{Dhillon2015}, and Fasttext \citep{Bojanowski2017}. These word vectors are usually referred to as \emph{distributional word vectors}, as their training methods rely on the \emph{distributional hypothesis} of semantics \citep{Firth1957}.

Recently, there has been interest in post-processing distributional word vectors to enrich their semantic content. The post-process procedures are usually performed in a lightweight fashion, i.e., without re-training word vectors on a text corpus. In one line of study, researchers used \emph{supervised methods} to enforce linguistic constraints (e.g., synonym relations) on word vectors \citep{Faruqui2015,Mrksic2016,Mrksic2017}, where the linguistic constraints are extracted from an external linguistic knowledge base such as WordNet \citep{Miller1995} and PPDB \citep{Pavlick2015}. In another line of study, researchers devised \emph{unsupervised methods} to post-process word vectors. Spectral-decomposition methods such as singular value decomposition (SVD) and principal component analysis (PCA) are usually used in this line of research \citep{Caron2001,Bullinaria2012,Turney2012,Levy2014,Levy2015,Mu2018}. The current paper is in line with the second, unsupervised, research direction. 

 Among different unsupervised word vector post-processing schemes, the \emph{all-but-the-top} approach \citep{Mu2018} is a prominent example. Empirically studying the latent features encoded by principal components (PCs) of distributional word vectors, \citet{Mu2018} found that the variances explained by the leading PCs ``encode the frequency of the word to a significant degree''. Since word frequencies are arguably unrelated to lexical semantics, they recommend removing such leading PCs from word vectors using a PCA reconstruction.

The current work advances the findings of \citet{Mu2018} and improves their post-processing scheme. Instead of discarding a fixed number of PCs, we \emph{softly} filter word vectors using \emph{matrix conceptors} \citep{Jaeger2014,Jaeger2017}, which characterize the linear space of those word vector features having high variances -- the features most contaminated by word frequencies according to \citet{Mu2018}. The proposed approach is mathematically simple and computationally efficient, as it is founded on elementary linear algebra. Besides these traits, it is also practically effective: using a standard set of lexical-level intrinsic evaluation tasks and a deep neural network-based dialogue state tracking task, we show that conceptor-based post-processing considerably enhances linguistic regularities captured by word vectors. A more detailed list of our contributions are:

\begin{enumerate}
    \item We propose an unsupervised algorithm that leverages Boolean operations of conceptors to post-process word vectors. The resulting word vectors achieve up to 18.86\% and 28.34\% improvement on the SimLex-999 and SimVerb-3500 dataset relative to the original word representations. 
    \item A closer look at the proposed algorithm reveals commonalities across several existing post-processing techniques for neural-based word vectors and pointwise mutual information (PMI) matrix based word vectors. Unlike the existing alternatives, the proposed approach is flexible enough to remove lexically-unrelated noise, while general-purpose enough to handle word vectors induced by different learning algorithms.
\end{enumerate}

The rest of the paper is organized as follows. We first briefly review the principal component nulling approach for unsupervised word vector post-processing introduced in \citep{Mu2018}, upon which our work is based. We then introduce our proposed approach, \emph{Conceptor Negation} (CN). Analytically, we reveal the links and differences between the CN approach and the existing alternatives. Finally, we showcase the effectiveness of the CN method with numerical experiments\footnote{Our codes are available at \url{https://github.com/liutianlin0121/Conceptor-Negation-WV}}.

\subsection{Notation}
We assume a collection of words $w \in V$, where $V$ is a vocabulary set. Each word $w \in V$ is embedded as a $n$ dimensional real valued vector $v_w \in \mathbb{R}^n$. An identity matrix will be denoted by $\mathbf{I}$. For a vector $v$, we denote \text{diag}($v$) as the diagonal matrix with $v$ on its diagonal. We write $[n] =  \{1, 2, \cdots, n\}$ for a positive integer $n$.

\section{Post-processing word vectors by PC removal}

This section is an overview of the all-but-the-top (ABTT) word vector post-processing approach introduced by \citet{Mu2018}. In brief, the ABTT approach is based on two key observations of distributional word vectors. First, using a PCA, \citet{Mu2018} revealed that word vectors are strongly influenced by a few leading principal components (PCs). Second, they provided an interpretation of such leading PCs: they empirically demonstrated a correlation between the variances explained by the leading PCs and word frequencies. Since word frequencies are arguably unrelated to lexical semantics, they recommend eliminating top PCs from word vectors via a PCA reconstruction. This method is described in Algorithm \ref{alg:abtt}.

\begin{algorithm}[htbp]
\SetKwInOut{Input}{Input}
\SetKwInOut{Output}{Output}
\Input{(i) $\{v_w \in \mathbb{R}^n : w \in V\}$: word vectors with a vocabulary $V$; (ii) $d$: the number of PCs to be removed.}
Center the word vectors: Let $\bar{v}_w \coloneqq v_w - \mu$ for all $w \in V$, where $\mu$ is the mean of the input word vectors. \\
Compute the first $d$ PCs $\{u_i \in \RR^n \}_{i \in [d]}$ of the column-wisely stacked centered word vectors $[\bar{v}_w]_{w \in V} \in \mathbb{R}^{n \times |V|}$ via a PCA.  \\
Process the word vectors: $\tilde{v}_w^{\text{ABTT}} \coloneqq  \bar{v}_w -  \sum_{i = 1}^d u_i^\top u_i \bar{v}_w, \forall w \in V$.  \\
\Output{$\{\tilde{v}_w^{\text{ABTT}}, w \in V\}$} 
\caption{The all-but-the-top (ABTT) algorithm for word vector post-processing.}
\label{alg:abtt}
\end{algorithm}

In practice, \citet{Mu2018} found that the improvements yielded by ABTT are particularly impressive for word similarity tasks. Here, we provide a  straightforward interpretation of the effects. Concretely, consider two arbitrary words $w_1$ and $w_2$ with word vectors $v_{w_1}$ and $v_{w_2}$. Without loss of generality, we assume $v_{w_1}$ and $v_{w_2}$ are normalized, i.e., $\| v_{w_1}\|_2 = \| v_{w_2}\|_2 = 1$. Given PCs $\{u_1, \cdots, u_n\}$ of the word vectors $\{v_w : w \in V\}$, we re-write $v_{w_1}$ and $v_{w_2}$  via linear combinations with respect to the basis $\{u_1, \cdots, u_n\}$: $v_{w_1} \coloneqq \sum_{i = 1}^n \beta_i u_i$ and $v_{w_2} \coloneqq \sum_{i = 1}^n \beta'_i u_i$, for some $\beta_i, \beta'_i \in \RR$ and for all $i \in [n]$. We see
\begin{eqnarray}
\text{cosine} (v_{w_1}, v_{w_2}) & \stackrel{( \ast)}{=} & v_{w_1}^\top v_{w_2}\label{eq:orig_cosine} \\
& = & \left (\sum_{i = 1}^n \beta_i u_i \right )^\top \left ( \sum_{i = 1}^n \beta'_i u_i \right ) \nonumber \\
& \stackrel{( \ast \ast)}{=} & \sum_{i = 1}^n \beta_i \beta'_i  \label{eq:cosine}
\end{eqnarray}
where $(\ast)$ holds because the word vectors were assumed to be normalized and $(\ast \ast)$ holds because $\{u_1, \cdots, u_n\}$ is an orthonormal basis of $\RR^n$. Via Equation \ref{eq:cosine}, the similarity between word $w_1$ and $w_2$ can be seen as the overall ``compatibility'' of their measurements $\beta_i$ and $\beta'_i$ with respect to each latent feature $u_i$. If leading PCs encode the word frequencies, removing the leading PCs, in theory, help the word vectors capture semantic similarities, and consequently improve the experiment results of word similarity tasks.

\section{Post-processing word vectors via Conceptor Negation}

Removing the leading PCs of word vectors using the ABTT algorithm described above is effective in practice, as seen in the elaborate experiments conducted by \citet{Mu2018}. However, the method comes with a potential limitation:  for each latent feature taking form as a PC of the word vectors,  ABTT either completely removes the feature or keeps it intact. For this reason, \citet{Khodak2018} argued that ABTT is liable either to not remove enough noise or to cause too much information loss. 

The objective of this paper is to address the limitations of ABTT. More concretely, we propose to use \emph{matrix conceptors}  \citep{Jaeger2017} to gate away variances explained by the leading PCs of word vectors. As will be seen later, the proposed \emph{Conceptor Negation} method removes noise in a ``softer'' manner when compared to  ABTT. We show that it shares the spirit of an eigenvalue weighting approach for PMI-based word vector post-processing. We proceed by providing the technical background of conceptors. 

\subsection{Conceptors}

Conceptors are a family of regularized identity maps introduced by \citet{Jaeger2014}. We present a sketch of conceptors by heavily re-using \citep{Jaeger2014,He2018} sometimes verbatim. In brief, a \textit{matrix conceptor} $C$ for some vector-valued random variable $x$ taking values in $\mathbb{R}^N$ is defined as a linear transformation that minimizes the following loss function.     
 \begin{align}
\label{eq:matconobj}
\mathbb{E} \left [ \|x - Cx \|_2^2 \right ]+\alpha^{-2}\|C\|_{F}^2
\end{align}
where $\alpha$ is a control parameter called \textit{aperture}, $\|\cdot\|_{2}$ is the $\ell_2$ norm, and $\|\cdot\|_{F}$ is the Frobenius norm. This optimization problem has a closed-form solution 
\begin{align}
\label{eq:conceptorsolution}
C = R(R+\alpha^{-2} \id )^{-1}
\end{align}
where $R=\mathbb{E}[xx^{\top}]$ and $\id$ are $N\times N$ matrices. If
$R = \Psi T \Psi^\top$ is the SVD of $R$, then the SVD of $C$ is given as $\Psi S \Psi^\top$, where the singular values $s_i$ of $C$ can be written in terms of the singular values $t_i$ of $R$: $s_i = t_i \slash (t_i + \alpha^{-2}) \in (0, 1)$ for $\alpha \in (0, \infty)$. In intuitive terms, $C$ is a soft projection matrix on the linear subspace where the samples of $x$ lie, such that for a vector $y$ in this subspace, $C$ acts like the identity: $Cy\approx y$, and when some $\epsilon$ orthogonal to the subspace is added to $y$, $C$ reconstructs $y$: $C(y+\epsilon)\approx y$. 

Moreover, operations that satisfy most laws of Boolean logic such as NOT $\neg$, OR $\vee$, and AND $\wedge$, can be defined on matrix conceptors. These operations all have interpretation on the data level, i.e., on the distribution of the random variable $x$ (details in \citep[Section 3.9]{Jaeger2014}).  Among these operations, the negation operation NOT $\neg$ is relevant for the current paper:
\begin{align}
\label{eq:bool}
\neg C \coloneqq \id -C.
\end{align}
Intuitively, the negated conceptor, $\neg C$, softly projects the data onto a linear subspace that can be roughly understood as the orthogonal complement of the subspace characterized by $C$. 

\subsection{Post-processing word vectors with Conceptor Negation}
This subsection explains how conceptors can be used to post-process word vectors. The intuition behind our approach is simple. Consider a random variable $x$ taking values on word vectors $\{v_w \in \mathbb{R}^n : w \in V\}$. We can estimate a conceptor $C$ that describes the distribution of $x$ using Equation \ref{eq:conceptorsolution}. Recall that  \citep{Mu2018} found that the directions with which $x$ has the highest variances encode word frequencies, which are unrelated to word semantics. To suppress such word-frequency related features, we can simply pass all word vectors through the negated conceptor $\neg C$, so that $\neg C$ dampens the directions with which $x$ has the highest variances. This simple method is summarized in Algorithm \ref{alg:cn}. 

\begin{algorithm}[htbp]
\SetKwInOut{Input}{Input}
\SetKwInOut{Output}{Output}
\Input{(i) $\{v_w \in \mathbb{R}^n : w \in V\}$: word vectors of a vocabulary $V$; (ii) $\alpha \in \RR$: a hyper-parameter}
Compute the conceptor $C$ from word vectors: $C = R(R+\alpha^{-2} \id )^{-1}$, where $R$ is estimated by $\frac{1}{|V|} \sum_{w} v_w v_w^\top $ \\
Compute $\neg C \coloneqq \id - C$ \\
Process the word vectors: $\tilde{v}_w^{\text{CN}} \coloneqq \neg C v_w, \forall w \in V$\\
\Output{$\{\tilde{v}_w^{\text{CN}} : w \in V\}$} 
\caption{The conceptor negation (CN) algorithm for word vector post-processing.}
\label{alg:cn}
\end{algorithm}

The hyper-parameter $\alpha$ of Algorithm \ref{alg:cn} governs the ``sharpness'' of the suppressing effects on word vectors employed by $\neg C$. Although in this work we are mostly interested in $\alpha \in (0, \infty)$, it is nonetheless illustrative to consider the extreme cases where $\alpha = 0$ or $\infty$: for $\alpha = 0$, $\neg C$ will be an identity matrix, meaning that word vectors will be kept intact; for $\alpha = \infty$, $\neg C$ will be a zero matrix, meaning that all word vectors will be nulled to zero vectors. The computational costs of the Algorithm \ref{alg:cn} are dominated by its step 1: one needs to calculate the matrix product $ R= \frac{1}{|V|} [v_{w}]_{w \in V} [v_{w}]_{w \in V}^{\top}$ for $[v_{w}]_{w \in V} \in \RR^{n \times |V|}$ being the matrix whose columns are word vectors. Since modern word vectors usually come with a vocabulary of some millions of words (e.g., Google News Word2Vec contains 3 million tokens), performing a matrix product on such large matrices $[v_{w}]_{w \in V}$ is computationally laborious. But considering that there are many uninteresting words in the vast vocabulary, we find it is empirically beneficial to only use a subset of the vocabulary, whose words are not too peculiar\footnote{This trick has also been used for ABTT by \citet{Mu2018} (personal communications).}. Specifically, borrowing the word list provided by \citet{Arora2017}\footnote{\url{https://github.com/PrincetonML/SIF/tree/master/auxiliary_data}}, we use the words that appear at least 200 times in a Wikipedia dump 2015 to estimate $R$. This greatly boosts the computation speed. Somewhat surprisingly, the trick also improves the performance of Algorithm \ref{alg:cn}. This might due to the higher quality of word vectors of common words compared with infrequent ones.

\section{Analytic comparison with other methods}

Since most of the existing unsupervised word vector post-processing methods are ultimately based on linear data transformations, we hypothesize that there should be commonalities between the methods. In this section, we show CN resembles ABTT in that both methods can be interpreted as ``spectral encode-decode processes''; when applied to word vectors induced by a pointwise mutual information (PMI) matrix, CN shares the spirit with the \emph{eigenvalue weighting} (EW) post-processing \citep{Caron2001,Levy2015}: they both assign weights on singular vectors of a PMI matrix. A key distinction of CN is that it does \emph{soft} noise removal (unlike ABTT) and that it is not restricted to post-processing PMI-matrix induced word vectors (unlike EW).

\subsection{Relation to ABTT} 
In this subsection, we reveal the connection between CN and ABTT. To do this, we will re-write the last step of both algorithms into different formats. For the convenience of comparison, throughout this section, we will assume that the word vectors $\{v_w\}_{v \in V}$ in Algorithm \ref{alg:abtt} and Algorithm \ref{alg:cn} possess a zero mean, although this is not a necessary requirement in general.

We first re-write the equation in step 3 of Algorithm \ref{alg:abtt}. We let $U$ be the matrix whose columns are the PCs estimated from the word vectors. Let $U_{:,1:d}$ be the first $d$ columns of $U$. It is clear that step 2 of Algorithm \ref{alg:abtt}, under the assumption that word vectors possess zero mean, can be re-written as 

\begin{align}
\tilde{v}_w^{\text{ABTT}} & \coloneqq \left ( I - U_{:,1:d} U_{:,1:d}^\top \right) v_w \nonumber \\
& = U \text{diag}([\underbrace{0, \cdots, 0}_{d~\text{copies~of~} 0}, 1, \cdots, 1]) U^\top v_w. \label{eq:abtt_rewrite}
\end{align}

Next, we re-write  step 3 of the Conceptor Negation (CN) method of algorithm \ref{alg:cn}. Note that for word vectors with zero mean, the estimation for $R$ is a (sample) covariance matrix of a random variable taking values as word vectors, and therefore the singular vectors of $R$ are PCs of word vectors. Letting $R = U\Sigma U^\top$ be the SVD of $R$, the equation in step 3 of Algorithm \ref{alg:cn} can be re-written via elementary linear algebraic operations: 
\begin{small}
\begin{align}
\tilde{v}_w^{\text{CN}} & \coloneqq \neg C  \bar{v}_w \nonumber \\
	&= \left ( \id - C \right ) v_w \nonumber \\
	&= \left ( \id - R (R + \alpha^{-2} I )^{-1} \right ) v_w \nonumber \\
	&= \left ( \id - U\Sigma U^\top (U\Sigma U^\top + \alpha^{-2} U U^\top )^{-1} \right ) v_w \nonumber \\
 	  & = \left ( \id - U \text{diag}([\frac{\sigma_{1}}{\sigma_{1}+\alpha^{-2}}, \cdots, \frac{\sigma_{n}}{\sigma_{n}+\alpha^{-2}}]) U^\top \right) v_w \nonumber \\
& = U \text{diag}([\frac{\alpha^{-2}}{\sigma_{1}+\alpha^{-2}}, \cdots, \frac{\alpha^{-2}}{\sigma_{n}+\alpha^{-2}}]) U^\top v_w, \label{eq:cn_rewrite}
\end{align}
\end{small}
where $\sigma_{1}, \cdots, \sigma_{n}$ are diagonal entries of $\Sigma$.

Examining Equations \ref{eq:abtt_rewrite} and \ref{eq:cn_rewrite}, we see ABTT and CN share some similarities. In particular, they both can be unified into ``spectral encode-decode processes,'' which contain the following three steps: 

\begin{enumerate}
\item \textbf{PC encoding}. Load word vectors on PCs by multiplying $U^\top$ with $v_w$.
\item \textbf{Variance gating}. Pass the PC-encoded data through the variance gating matrices $\text{diag}([0, \cdots, 0, 1, \cdots, 1])$ and $\text{diag}([\frac{\alpha^{-2}}{\sigma_{1}+\alpha^{-2}}, \cdots, \frac{\alpha^{-2}}{\sigma_{n}+\alpha^{-2}}])$ respectively for ABTT and CN. 
\item \textbf{PC decoding}. Transform the data back to the usual coordinates using the matrix $U$. 
\end{enumerate} 

\begin{figure}[htbp]
\centering
\includegraphics[width = 0.4\textwidth]{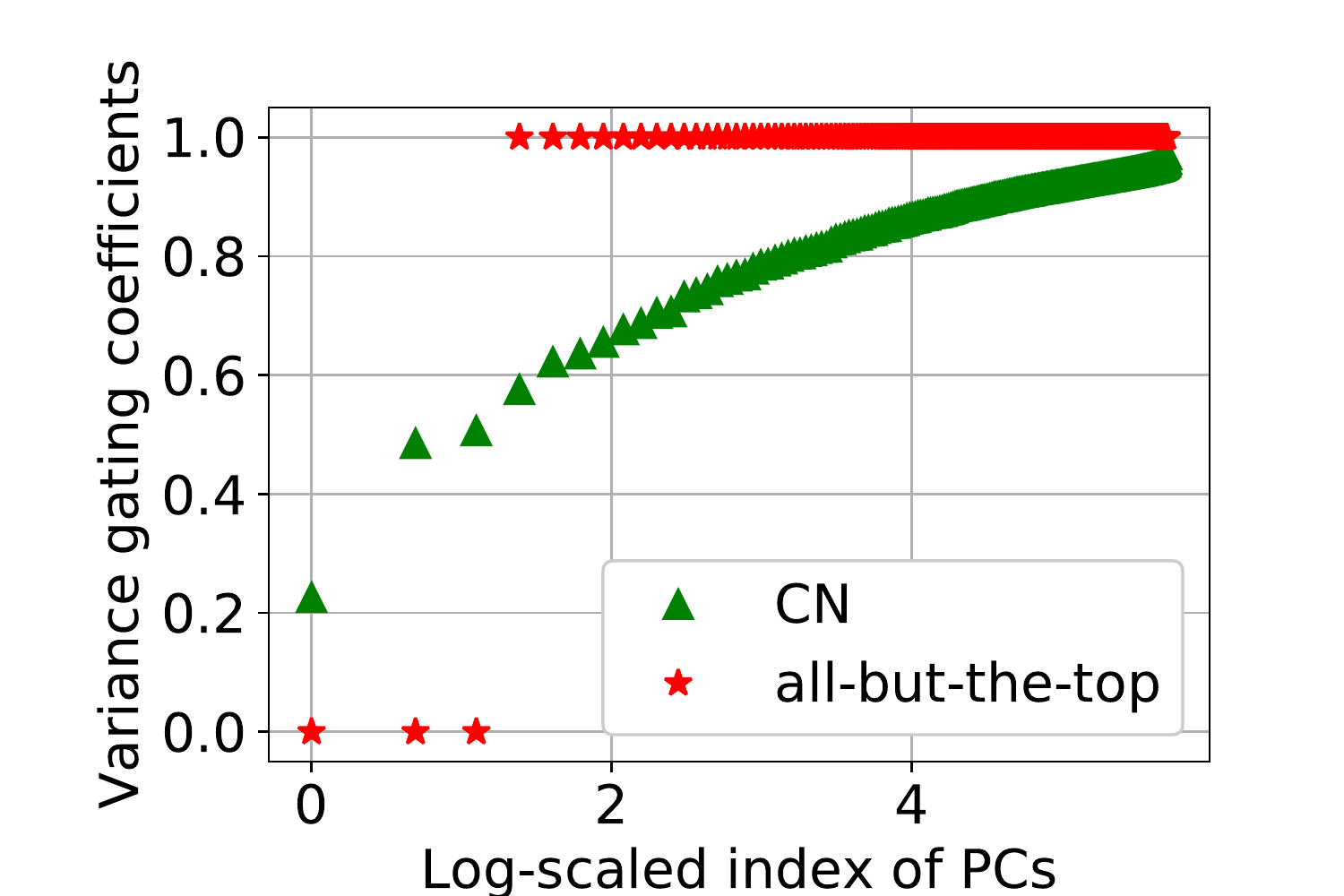}
\caption{The variance gating coefficients of ABTT and CN for Word2Vec. Hyper-parameters: $d = 3$ for ABTT and $\alpha = 2$ for CN.}
\label{fig:gatingCoeff}
\end{figure}

With the above encode-decode interpretation, we see CN differ from ABTT  is its variance gating step. In particular, ABTT does a \emph{hard} gating, in the sense that the diagonal entries of the variance gating matrix (call them variance gating coefficients) take values in the set $\{0,1\}$. The CN approach, on the other hand, does a softer gating as the entries take values in $(0,1)$:
\[ 0 < \frac{\alpha^{-2}}{\sigma_{i}+\alpha^{-2}} \leq \frac{\alpha^{-2}}{\sigma_{j}+\alpha^{-2}} < 1,\]  for all $1 \leq i < j \leq n$ and $\alpha \in (0, \infty)$. To illustrate the gating effects, we plot the variance gating coefficients for ABTT and CN for Word2Vec in Figure \ref{fig:gatingCoeff}.

\subsection{Relation with eigenvalue weighting}

We relate the conceptor approach to the eigenvalue weighting approach for post-processing PMI-based word vectors. This effort is in line with the ongoing research in the NLP community that envisages a connection between ``neural word embedding'' and PMI-matrix factorization based word embedding \citep{Levy2014,Pennington2014,Levy2015}.

In the PMI approach for word association modeling, for each word $w$ and each context (i.e., sequences of words) $q$, the PMI matrix $M$ assigns a value for the pair $(w,q)$: $M(w, q) = \log \frac{\mathbb{P}( w, q)}{\mathbb{P}(w) \mathbb{P} (q)}$. In practical NLP tasks, the sets of words and contexts tend to be large, and therefore, directly working with $M$ is inconvenient. To lift the problem, one way is to perform a truncated SVD on $M$, factorizing $M$ into the product of three smaller matrices $M \approx  \Theta_{:, 1:n} D_{1:n, 1:n} \Gamma_{:, 1:n}^\top$, where $\Theta_{:, 1:n}$ is the first $n$ left singular vectors of the matrix $M$, $D_{1:n, 1:n}$ is the diagonal matrix containing $n$ leading singular values of $M$, and $\Gamma_{:, 1:n}$ are the first $n$ right singular vectors of the matrix $M$. A generic way to induce word vectors from $M$ is to let \[ E \coloneqq \Theta_{:, 1:n} D_{1:n, 1:n} \in \RR^{|V| \times n}, \] which is a matrix containing word vectors as rows. Coined by \citet{Levy2015}, the term \emph{eigenvalue weighting}\footnote{It seems to us that a term ``singular value weighting'' is more appropriate because the weighting is based on singular values of a PMI matrix $M$ but not eigenvalues of $M$. The term ``eigenvalue'' is relevant here only because the singular values of $M$ are also the square roots of eigenvalues of $M^\top M$.} (EW) refers to a post-processing technique for PMI-matrix-induced word vectors. This technique has its root in Latent Semantic Analysis (LSA): \citet{Caron2001} first propose to define the post-processed version of $E$ as
\[ \tilde{E}^{\text{EW}} \coloneqq \Theta_{:, 1:n} D_{1:n, 1:n}^p,\] 
where $p$ is the \emph{weighting exponent} determining the relative weights assigned to each singular vector of $\Theta_{:, 1:n}$. While an optimal $p$ depends on specific task demands, previous research suggests that $p < 1$ is generally preferred, i.e., the contributions of the initial singular vectors of $M$ should be suppressed. For instance, $p = 0, 0.25$, and $0.5$ are recommended in \citep{Caron2001,Bullinaria2012,Levy2015}. \citet{Bullinaria2012} argue that the initial singular vectors of $M$ tend to be contaminated most by aspects other than lexical semantics.

We now show that applying CN on the PMI-matrix-based word embedding $E \coloneqq \Theta_{:, 1:n} D_{1:n, 1:n}$ has a tantamount effect with ``suppressing initial singular vectors'' of EW. Acting the negated $\neg C$ on word vectors of $E$ (i.e., rows of $E$), we get the post-processed word vectors as rows of the $\tilde{E}^{\text{CN}}$:

\begin{small}
\begin{align*}
\tilde{E}^{\text{CN}} & \coloneqq  (\neg C E^\top)^\top  \\
& = E (I - R (R +  \alpha^{-2} I)^{-1})  \\
& = E (I - \frac{1}{|V|} E^\top E (\frac{1}{|V|} E^\top E + \alpha^{-2} I)^{-1})  \\
& = \Theta_{:,1:n} D_{1:n, 1:n} (I - \frac{1}{|V|}  D_{1:n, 1:n}^2 ( \frac{1}{|V|} D_{1:n, 1:n}^2 +  \alpha^{-2} I)^{-1})  \\
& =  \Theta_{:,1:n} D_{1:n, 1:n} \text{diag}([ \frac{|V| \cdot \alpha^{-2}}{  \lambda_1^2 + |V| \cdot \alpha^{-2}}, \cdots,  \frac{|V| \cdot \alpha^{-2}}{ \lambda_n^2 + |V| \cdot \alpha^{-2}} ])\label{eq:cnPmi} 
\end{align*}
\end{small}

Since 
\[ 0 < \frac{|V| \cdot \alpha^{-2}}{  \lambda_i^2 + |V| \cdot \alpha^{-2}}  \leq \frac{|V| \cdot \alpha^{-2}}{ \lambda_j^2 + |V| \cdot \alpha^{-2} } < 1,\]  for all $1 \leq i < j \leq n$ and $\alpha \in (0, \infty)$, these weights suppress the contribution of the initial singular vectors, similar to what has been done in EW.

  \section{Experiments}
  
 We evaluate the post-processed word vectors on a variety of lexical-level intrinsic tasks and a down-stream deep learning task. We use the publicly available pre-trained Google News Word2Vec \citep{Mikolov2013}\footnote{\url{https://code.google.com/archive/p/word2vec/}} and Common Crawl GloVe\footnote{\url{https://nlp.stanford.edu/projects/glove/}} \citep{Pennington2014} to perform lexical-level experiments. For CN, we fix $\alpha = 2$ for Word2Vec and GloVe throughout the experiments\footnote{Analytical optimization methods for the aperture $\alpha$ are available from \citep{Jaeger2014}, remaining to be connected with the word vector post-processing scheme in the future.}. For ABTT, we set $d = 3$ for Word2Vec and $d = 2$ for GloVe, as what has been suggested by \citet{Mu2018}.  
 
\paragraph{Word similarity} We test the performance of CN on seven benchmarks that have been widely used to measure word similarity: the RG65 \citep{Rubenstein1965},  the WordSim-353 (WS) \citep{Finkelstein2002},  the rare-words (RW) \citep{Luong2013}, the MEN dataset \citep{Bruni2014}, the MTurk \citep{Radinsky2011}, the SimLex-999 (SimLex) \citep{Hill2015}, and the SimVerb-3500 \citep{Gerz2016}. To evaluate the word similarity, we calculate the cosine distance between vectors of two words using Equation \ref{eq:orig_cosine}. We report the Spearman's rank correlation coefficient \citep{Myers1995} of the estimated rankings against the rankings by humans in Table \ref{tb:similarity}. We see that the proposed CN method consistently outperforms the original word embedding (orig.) and the post-processed word embedding by ABTT for most of the benchmarks.

\begin{table}[htbp] 
\scalebox{0.85}{
\begin{tabular}{r  c c c c c c}
\toprule
\multirow{2}{*}{} & \multicolumn{3}{c}{WORD2VEC} & \multicolumn{3}{c}{GLOVE} \\  \cmidrule(r){2-4} \cmidrule(r){5-7}
                  &  orig.      &  ABTT &  CN     &  orig.    &  ABTT &  CN  \\ \midrule
 RG65  &     76.08 &       78.34  & \bf 78.92  &  76.96   &  74.36   &  \bf 78.40       \\ 
 WS    &     68.29 &       69.05 &\bf 69.30 &  73.79    &   76.79 & \bf 79.08  \\ 
 RW    &     53.74 &      54.33 & \bf 58.04 & 46.41   &   52.04    & \bf 58.98   \\ 
 MEN   &     78.20 &     \bf 79.08 & 78.67 & 80.49   &   81.78  & \bf 83.38      \\ 
 MTurk &     68.23 &     \bf 69.35 & 66.81 & 69.29  &    70.85  & \bf 71.07  \\ 
 SimLex  &   44.20 &    45.10   &  \bf 46.82 & 40.83 &  44.97     & \bf 48.53    \\ 
 SimVerb &   36.35 &   36.50   & \bf 38.30 &  28.33   &   32.23    & \bf 36.36    \\ \bottomrule
\end{tabular}}
\caption{Post-processing results (Spearman's rank correlation coefficient $\times$ 100) under seven word similarity benchmarks. The baseline results (orig. and ABTT) are collected from \citep{Mu2018}. }
\label{tb:similarity}
\end{table}
The improvement of results by CN are particularly impressive for two ``modern'' word similarity benchmarks SimLex and SimVerb -- these two benchmarks carefully distinguish \emph{genuine word similarity} from \emph{conceptual association} \citep{Hill2015}. For instance, \texttt{coffee} is associated with \texttt{cup} but by no means similar to \texttt{cup}, a confusion often made by earlier benchmarks. In particular, SimLex has been heavily used to evaluate word vectors yielded by \emph{supervised} word vector fine-tuning algorithms, which perform gradient descent on word vectors with respect to linguistic constraints such as synonym and antonym relationships of words extracted from WordNet and/or PPDB. When compared to a recent supervised approach of counter-fitting. Our results on SimLex are comparable to those reported by \citet{Mrksic2016}, as shown in Table \ref{tb:similarity_supervised}. 

\begin{table}[htbp] 
\scalebox{0.8}{
\begin{tabular}{l  c  c c } \toprule
   &Post-processing method & WORD2VEC & GLOVE    \\ \midrule
\multirow{3}{*}{supervised} & Counter-Fitting + syn. &      0.45   &  0.46       \\ 
 & Counter-Fitting + ant. &      0.33   & 0.43        \\ 
 & Counter-Fitting + syn. + ant. &  \bf 0.47 & \bf 0.50 \\ \hdashline
unsupervised  & CN & \bf 0.47 &  0.49 \\ \bottomrule
\end{tabular}}
\caption{Comparing the testing results (Spearman's rank correlation coefficient) on SimLex with those of Counter-Fitting approach (results collected from \citep[Table 2]{Mrksic2016} and \citep[Table 3]{Mrksic2017}). The linguistic constraints for Counter-Fitting are synonym (syn.) and/or antonym (ant.) relationships extracted from English PPDB.}
\label{tb:similarity_supervised}
\end{table}

\paragraph{Semantic Textual Similarity} In this subsection, we showcase the effectiveness of the proposed post-processing method using semantic textual similarity (STS) benchmarks, which are designed to test the semantic similarities of sentences. We use 2012-2015 SemEval STS tasks \citep{Agirre2012,Agirre2013,Agirre2014,Agirre2015} and the 2012 SemEval Semantic Related task (SICK) \citep{Marco2014}.

Concretely, for each pair of sentences, $s_1$ and $s_2$, we computed $v_{s_1}$ and $v_{s_2}$ by averaging their constituent word vectors. We then calculated the cosine distance between two sentence vectors $v_{s_1}$ and $v_{s_2}$. This naive method has been shown to be a strong baseline for STS tasks \citep{Wieting2016}. As in \citet{Agirre2012}, we used Pearson correlation of the estimated rankings of sentence similarity against the rankings by humans to assess model performance. 

In Table \ref{tb:sts:detail}, we report the average result for the STS tasks each year (detailed results are in the supplemental material). Again, our CN method consistently outperforms the alternatives.

\begin{table}[htbp]
\scalebox{0.85}{
\centering
\begin{tabular}{ r c c c c c c}
\hline
\multirow{2}{*}{}     & \multicolumn{3}{c}{WORD2VEC}                                    & \multicolumn{3}{c}{GLOVE}                                       \\   \cmidrule(r){2-4} \cmidrule(r){5-7}
                      	& orig.                    	& ABTT                 & CN              	& orig.                      	& ABTT              	& CN                 \\ \midrule
STS 2012		& 57.22		        	& \bf 57.67		&  54.31	 	& 48.27			&54.06			&\bf 54.38	\\
STS 2013		&56.81		    	& 57.98		& \bf 	59.17	&44.83			& 51.71			& \bf 55.51 	\\
STS 2014		&	62.89.   		&  63.30		& \bf  66.22		& 51.11			& 59.23			& \bf 62.66		\\
STS 2015		&62.74			& 63.35		&  \bf	 67.15	&47.23			& 57.29			& \bf 	63.74\\
SICK                  & 70.10                      & {70.20}     & \bf 72.71                 & 65.14                      & \bf{67.85}       		& 66.42               \\ 
\bottomrule
\end{tabular}
}
\caption{Post-processing results ($\times $100) on the semantic textual similarity tasks. The baseline results (orig. and ABTT) are collected from \citep{Mu2018}.}
\label{tb:sts:detail}
\end{table}

\paragraph{Concept Categorization} In the concept categorization task, we used $k$-means to cluster words into concept categories based on their vector representations (for example, ``bear'' and ``cat'' belong to the concept category of animals). We use three standard datasets: (i) a rather small dataset ESSLLI 2008 \citep{Baroni2008} that contains 44 concepts in 9 categories; (ii) the Almuhareb-Poesio (AP) \citep{Poesio2005}, which contains 402 concepts divided into 21 categories; and (iii) the BM dataset \citep{Battig1969} that 5321 concepts divided into 56 categories. Note that the datasets of ESSLLI, AP, and BM are increasingly challenging for clustering algorithms, due to the increasing numbers of words and categories.

Following \citep{Baroni2014,Schnabel2015,Mu2018}, we used ``purity'' of clusters \citep[Section 16.4]{Manning2008} as the evaluation criterion. That the results of $k$-means heavily depend on two hyper-parameters: (i) the number of clusters and (ii) the initial centroids of clusters. We follow previous research \citep{Baroni2014,Schnabel2015,Mu2018} to set $k$ as the ground-truth number of categories. The settings of the initial centroids of clusters, however, are less well-documented in previous work -- it is not clear how many initial centroids have been sampled, or if different centroids have been sampled at all. To avoid the influences of initial centroids in $k$-means (which are particularly undesirable for this case because word vectors live in $\RR^{300}$), in this work, we simply fixed the initial centroids as the average of original, ABTT-processed, and CN-processed word vectors respectively from ground-truth categories. This initialization is fair because all post-processing methods make use of the ground-truth information equally, similar to the usage of the ground-truth numbers of clusters. We report the experiment results in Table \ref{tb:categorization}.

\begin{table}[htbp]
\scalebox{0.85}{
\begin{tabular}{ r c c c c c c }
\toprule
\multirow{2}{*}{} & \multicolumn{3}{c}{WORD2VEC} & \multicolumn{3}{c}{GLOVE} \\ \cmidrule(r){2-4} \cmidrule(r){5-7}
                  & orig.      & ABTT       & CN.    	 & orig.   	 	& ABTT  & CN  \\ \midrule
ESSLLI &100.0 &     100.0 &    100.0 	&  100.0 		&  100.0     & 100.0\\ 
AP 		& 87.28	 &  88.3  & \bf 89.31 	&   86.43		&   87.19  & \bf 90.95 \\ 
BM    	 & 58.15 	&    59.24 & \bf 60.19   	&  65.34 	& 67.35  & \bf 67.63  \\ 
\bottomrule
\end{tabular}
}
\caption{Purity ($\times$ 100) of the clusters in concept categorization task with fixed centroids.}
\label{tb:categorization}
\end{table}

The performance of the proposed methods and the baseline methods performed equally well for the smallest dataset ESSLLI. As the dataset got larger, the results differed and the proposed CN approach outperformed the baselines. 

\paragraph{A Downstream NLP task: Neural Belief Tracker} The experiments we have reported so far are all intrinsic lexical evaluation benchmarks. Only evaluating the post-processed word vectors using these benchmarks, however, invites an obvious critique: the success of intrinsic evaluation tasks may not transfer to downstream NLP tasks, as suggested by previous research \citep{Schnabel2015}. Indeed, when supervised learning tasks are performed, the post-processing methods such as ABTT and CN can \emph{in principle} be absorbed into a classifier such as a neural network. Nevertheless, good initialization for classifiers is crucial. We hypothesize that the post-processed word vectors serve as a good initialization for those downstream NLP tasks that semantic knowledge contained in word vectors is needed. 

To validate this hypothesis, we conducted an experiment using Neural Belief Tracker (NBT), a deep neural network based dialogue state tracking (DST) model \citep{Mrksic2017,Mrksic2018}. As a concrete example to illustrate the purpose of the task, consider a dialogue system designed to help users find restaurants. When a user wants to find a Sushi restaurant, the system is expected to know that Japanese restaurants have a higher probability to be a good recommendation than Italian restaurants or Thai restaurants. Word vectors are important for this task because NBT needs to absorb useful semantic knowledge from word vectors using a neural network. 

In our experiment with NBT, we used the model specified in \citep{Mrksic2018} with default hyper-parameter settings\footnote{\url{https://github.com/nmrksic/neural-belief-tracker}}. We report the \emph{goal accuracy}, a default DST performance measure, defined as the proportion of dialogue turns where all the user's search goal constraints match with the model predictions. The test data was  Wizard-of-Oz (WOZ) 2.0 \citep{Wen2017}, where the goal constraints of users were divided into three domains: \emph{food}, \emph{price range}, and \emph{area}. The experiment results are reported in Table \ref{tb:NBT}.

\begin{table}[h!]
\scalebox{0.85}{
\begin{tabular}{ r c c c c c c }
\toprule
\multirow{2}{*}{} & \multicolumn{3}{c}{WORD2VEC} & \multicolumn{3}{c}{GLOVE} \\ \cmidrule(r){2-4} \cmidrule(r){5-7}
                 	& orig.      & ABTT       	& CN.    	 	& orig.   	 	& ABTT  		& CN  \\  \midrule
Food 		& 48.6 	&  \bf 84.7 	&  78.5 		& 86.4 		&  83.7   		&\bf  88.8\\ 
Price range 	& 90.2	&  88.1  		& \bf 92.2 		& 91.0		&  93.9  		& \bf 94.7 \\ 
Area    	 	& 83.1 	&  82.4 		& \bf 86.1   	& 93.5 		& \bf 94.9  	&  93.7  \\  \hdashline
Average    	& 74.0 	&  85.1 		& \bf 85.6   	& 90.3 		& 90.8  		&  \bf 92.4  \\ 
\bottomrule
\end{tabular}
}
\caption{The goal accuracy of food, price range, and area.}
\label{tb:NBT}
\end{table}

\paragraph{Further discussions}
Besides the NBT task, we have also tested ABTT and CN methods on other downstream NLP tasks such as text classification (not reported). We found that ABTT and CN yield equivalent results in such tasks. One explanation is that the ABTT and CN post-processed word vectors are different only up to a small perturbation. With a sufficient amount of training data and an appropriate regularization method, a neural network should generalize over such a perturbation. With a relatively small training data (e.g., the 600 dialogues for training NBT task), however, we found that word vectors as initializations matters, and in such cases, CN post-processed word vectors yield favorable results. Another interesting finding is that having tested ABTT and CN on Fasttext \citep{Bojanowski2017}, we found that neither post-processing method provides visible gain. We hypothesize that this might be because Fasttext includes subword (character-level) information in its word representation during training, which suppresses the word frequency features contained in word vectors. It remains for future work to validate this hypothesis.
\section{Conclusion}

We propose a simple yet effective method for post-processing word vectors via the negation operation of conceptors. With a battery of intrinsic evaluation tasks and a down-stream deep-learning empowered dialogue state tracking task, the proposed method enhances linguistic regularities captured by word vectors and consistently improves performance over existing alternatives.

There are several possibilities for future work. We envisage that the logical operations and abstract ordering admitted by conceptors can be used in other NLP tasks. As concrete examples, the AND $\wedge$ operation can be potentially applied to induce and fine-tune bi-lingual word vectors, by mapping word representations of individual languages into a shared linear space; the OR $\vee$ together with NOT $\neg$ operation can be used to study the vector representations of polysemous words, by joining and deleting sense-specific vector representations of words; the abstraction ordering $\leq$ is a natural tool to study graded lexical entailment of words.

\paragraph{Acknowledgement} We appreciate the anonymous reviewers for their constructive comments. We thank Xu He, Jordan Rodu, and Daphne Ippolito, and Chris Callison-Burch for helpful discussions. 

\bibliography{nlp_progress.bib}
\bibliographystyle{aaai}

%APPENDIX
\newpage
\onecolumn
%\vfill
\section*{Appendix}

\section{Detailed experiments in Semantic Textual Similarity (STS) tasks}
In the main body of the paper we have reported the average results for STS tasks by year. A detailed list the STS tasks can be found in Table \ref{tb:sts:tasks} and can be downloaded from \url{http://ixa2.si.ehu.es/stswiki/index.php/STSbenchmark}.

\begin{table}[h]
\centering
\begin{tabular}{ r c c c}
\toprule
STS 2012 & STS 2013 & STS 2014 & STS 2015  \\
\midrule
MSRpar & FNWN & deft forum & anwsers-forums \\
MSRvid & OnWN & deft news & answers-students \\
OnWN & headlines & headline & belief \\
SMTeuroparl &  & images & headline\\
SMTnews & & OnWN & images\\
& &   tweet news &  \\
\bottomrule
\end{tabular}
\caption{STS tasks in year 2012 - 2015. Note that tasks with shared names in different years are different tasks.}
\label{tb:sts:tasks}
\end{table}

We report the detailed experiment results for the above STS tasks in Table \ref{tb:sts:detail}.

\begin{table}[h]
\centering
\begin{tabular}{ r c c c c c c}
\hline
\multirow{2}{*}{} & \multicolumn{3}{c}{WORD2VEC} & \multicolumn{3}{c}{GLOVE} \\  \cmidrule(r){2-4} \cmidrule(r){5-7}
                      & orig.                      	& ABTT                 & CN              		& orig.                      & ABTT              & CN                 \\ \hline
MSRpar           & 42.12                      & \bf 43.85    	    & 40.30                  	& \textbf{44.54}             & 44.09            & 41.19	                  \\ 
MSRvid           & 72.07                      & 72.16     		    & \bf 75.22                 	& 64.47                      & \bf{68.05}        & 62.50	              \\ 
OnWN             & 69.38                      & 69.48        &   \bf  70.82          & 53.07                      & {65.67}        & \bf 67.96	              \\ 
SMTeuroparl      & 53.15                      & \bf 54.32         & 35.14             & 41.74                      & {45.28}       &\bf 52.58	               \\ 
SMTnews          & 49.37             & 48.53            & \bf 50.08                   & 37.54                      & {47.22}       & \bf 47.69               \\ \hdashline
STS 2012		& 57.22		        & 57.67			& \bf 54.31		&	  48.27				&	54.06			&\bf 54.38	\\
\midrule
FNWN             & 40.70                      & 41.96        & \bf 43.99              & 37.54                      & {39.34}       & \bf 42.07	               \\ 
OnWN             & 67.87                      & 68.17         & \bf 68.76            & 47.22                      & \bf{58.60}     & 57.45	                 \\ 
headlines        & 61.88 		     & 63.81 		&  \bf 64.78 		& 49.73.     	& {57.20} 	& \bf 67.00  \\ \hdashline
STS 2013		&56.81			        & 57.98		& \bf 59.17			&44.83					& 51.71				& \bf 55.51	\\
\midrule
OnWN             & 74.61                      & 74.78         & \bf 75.08             & 57.41                      & \bf{67.56}        & 66.43              \\ 
deft-forum       & 32.19                      & 33.26          & \bf 42.80            & 21.55                      & {29.39}         & \bf 37.57             \\ 
deft-news        & 66.83             & \bf 65.96                 & 65.57              & 65.14                      & \bf{71.45}        & 69.08	              \\ 
headlines        & 58.01                      & 59.58       & \bf 61.09               & 47.05                      & {52.60}        & \bf 61.71	              \\ 
images           & 73.75                      & 74.17       & \bf 78.24               & 57.22                      &\bf {68.28}         & 65.81             \\ 
tweet-news       & 71.92                      & 72.07       & \bf 74.55               & 58.32                      & {66.13}       & \bf 75.37	               \\ \hdashline
STS 2014		&	62.89		        &  63.30			& \bf 66.22		& 51.11					&	59.23		& \bf 62.66		\\
\midrule
forum    & 46.35                      & 46.80        & \bf 53.66              & 30.02                      & {39.86}         & \bf 48.62             \\ 
students & 68.07             & 67.99             & \bf 71.45                  & 49.20                      & {62.38}         & \bf 69.68             \\ 
belief           & 59.72                      & 60.42       & \bf 61.29              & 44.05                      & {57.68}              & \bf 59.77	        \\ 
headlines        & 61.47                      & 63.45      &\bf 68.88                & 46.22                      & {53.31}         & \bf 69.18	            \\ 
images           & 78.09                      & 78.08     & \bf 80.48                & 66.63                      & \bf{73.20}         & 71.43	           \\ \hdashline
STS 2015		&62.74			        & 63.35			&  \bf 67.15		&	47.23				& 57.29			& \bf 63.74	\\
\midrule
SICK                  & 70.10                      & {70.20}     & \bf 72.71                 & 65.14                      & \bf{67.85}       & 66.42               \\ 
\bottomrule
\end{tabular}
\caption{Before-After results (x100) on the semantic textual similarity tasks.}
\label{tb:sts:detail}
\end{table}

\end{document}